\documentclass[preprint,12pt,square,sort,comma,numbers]{elsarticle}

\usepackage{amssymb}

\usepackage[utf8]{inputenc} 
\usepackage[T1]{fontenc}    
\usepackage{hyperref}       
\usepackage{url}            
\usepackage{booktabs}       
\usepackage{amsfonts}       
\usepackage{nicefrac}       
\usepackage{microtype}      

\usepackage{amsmath}
\usepackage{tikz}
\usetikzlibrary{matrix}

\usepackage{float}
\usepackage[caption = false]{subfig}

\usepackage{graphicx, multicol}

\newcommand{\eq}{\begin{equation}}
\newcommand{\eqx}{\end{equation}}
\newcommand{\eqn}{\begin{eqnarray}}
\newcommand{\eqnx}{\end{eqnarray}}
\newcommand{\f}[2]{\frac{#1}{#2}}

\newcommand{\lm}{\lambda}

\newcommand{\qq}{\quad\quad}

\newcommand{\cor}[1]{\left\langle{#1}\right\rangle}

\usepackage[colorinlistoftodos]{todonotes}

\begin{document}

\begin{frontmatter}



\title{Complexity for deep neural networks\\ and other characteristics of deep feature representations}


\author[1]{Romuald A. Janik}

\author[2]{Przemek Witaszczyk}

\address[1]{Institute of Theoretical Physics\\
  Jagiellonian University\\
  Kraków, Poland \\
  \texttt{romuald.janik@gmail.com}
}

\address[2]{Institute of Theoretical Physics\\
  Jagiellonian University\\
  Kraków, Poland \\
  \texttt{przemek.witaszczyk@cern.ch}}

\begin{abstract}
We define a notion of complexity, 
which quantifies the nonlinearity of the computation of a neural network, as well as a complementary measure of the effective dimension of feature representations. We investigate these observables
both for trained networks for various datasets as well as explore their dynamics during training, uncovering in particular power law scaling.
These observables can be understood in a dual way as uncovering hidden internal structure of the datasets themselves as a function of scale or depth.
The entropic character of the proposed notion of complexity should allow to transfer modes of analysis from neuroscience and statistical physics to the domain of artificial neural networks.
The introduced observables can be applied without any change to the analysis of biological neuronal systems.
\end{abstract}




\end{frontmatter}



\section{Introduction}
\label{s.intro}

The goal of the present paper is to introduce characterizations of complexity
tailored to Deep Neural Networks. 
In Machine Learning, typically model complexity is associated with the number of model parameters
and thus with its potential expressivity. In this work we would like to focus, in contrast, on a notion
which captures the complexity of the specific computational task, so that the
same neural network architecture with the same total number of parameters could
yield differing complexity depending on the dataset on which it was trained.


The motivation for our specific version of complexity comes from the notion of \emph{circuit complexity} in Quantum Information and Physics,
although our definition is not a direct translation of that concept to neural networks.
Another aspect that we would like to emphasize is that the entropic character of our definition lends itself on the one hand
to considerations of efficiency or effectiveness and on the other hand is a direct counterpart of a method of analysis of
ensembles spiking neurons in neuroscience~\cite{IsingNeurons}.

Another goal for the considerations of the present paper is that the proposed definitions
could be used to study information processing and feature representations in a deep neural network in a \emph{layer-wise} manner.
Intuitively, a (convolutional) deep neural network builds up first local representations (edges/textures) then
going further in depth, the features become more nonlocal and more higher-level culminating in the final classification score.
The deep convolutional neural network can thus be understood as a highly nontrivial nonlinear \emph{goal directed} analog of a Fourier/wavelet transform of the original image.
Similar flavour of analysis is ubiquitous in Physics, where one looks at a system at a progression of scales. This occurs e.g.
in the context of (Wilsonian) Renormalization Group, tensor networks or holography (see more in section~\ref{s.relatedwork}).
Thus it is very interesting to introduce \emph{quantitative} characteristics of feature representations at a given scale/layer depth and study their variation as  a function of depth.

Let us note that this opens up also a quite different theoretical perspective of characterizing, in a nontrivial way, the internal structure at a given scale of a specific dataset or learning task using the trained deep neural network as \emph{a tool} instead of a classifier.

In this respect, as our notion of complexity captures only the nonlinear character of information processing of a given layer,
we also introduce a complementary measure -- \emph{the effective dimension}, which is sensitive to the details of neurons working in the linear regime. 

Naturally, once such notions are introduced, it is interesting to study how these characteristics evolve during training, which may provide both qualitative insight as well as quantitative data for the understanding of the dynamics of deep neural network training.

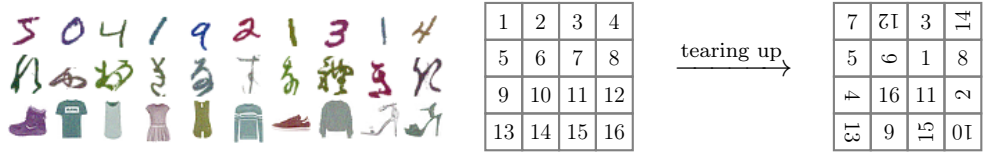
\begin{figure}[t]
\centering
\begin{tabular}{cccc}
\raisebox{-.4\height}{
\includegraphics[width=0.43\textwidth]{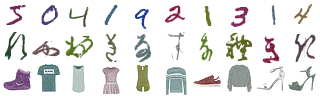}} 

\begin{tabular}{c}
\begin{tikzpicture}[thick,scale=0.7, every node/.style={scale=0.7}]
\matrix[matrix of nodes,nodes={draw=gray, anchor=center, minimum size=.7cm}, column sep=-\pgflinewidth, row sep=-\pgflinewidth] (A) {
1 & 2 & 3 & 4 \\
5 & 6 & 7 & 8 \\
9 & 10 & 11 & 12 \\
13 & 14 & 15 & 16\\};
\end{tikzpicture}
\end{tabular} 

\hspace{-0.3cm}

\raisebox{0.25cm}{\begin{tabular}{c}
$\xrightarrow{\text{tearing up}}$
\end{tabular}}

\hspace{-0.3cm}

\begin{tabular}{c}
\begin{tikzpicture}[thick,scale=0.7, every node/.style={scale=0.7}]
\matrix[matrix of nodes,nodes={draw=gray, anchor=center, minimum size=.7cm}, column sep=-\pgflinewidth, row sep=-\pgflinewidth] (A) {
7 & \rotatebox{180}{12} & 3 & \rotatebox{90}{14} \\
5 &  \rotatebox{90}{6} & 1 & 8 \\
\rotatebox{270}{4} & 16 & 11 & \rotatebox{90}{2} \\
\rotatebox{270}{13} & 9 & \rotatebox{90}{15} & \rotatebox{180}{10}\\};
\end{tikzpicture}
\end{tabular} 

\end{tabular} 

\caption{CMNIST, CKMNIST and CFashionMNIST datasets (left).
Teared up version (right). See \ref{s.datasets} for details.}
\label{fig.datasets}
\end{figure}

In order to investigate how the introduced observables differ once one changes the difficulty and type of learning task while keeping the underlying neural network
unchanged, we utilize a range of datasets.
Since our base network is geared for CIFAR-10, we introduce as drop-in replacements
coloured variants of the classical MNIST, KMNIST and FashionMNIST datasets as examples of easier tasks, as well as novel teared-up
versions of all these datasets which serve to increase the difficulty of the classification tasks (see Fig.~\ref{fig.datasets} and \ref{s.datasets}). 
We also investigate a memorization task, namely CIFAR-10 with random labels~\cite{memorization}. 

The plan of the paper is as follows. In section~2, we provide further motivations as well as discuss the similarities and differences with related work. In section~3 we briefly review our treatment of convolutional layers for the subsequent computation of the introduced observables. The following two sections contain our key new definitions -- \emph{complexity} and \emph{effective dimension}. We then proceed to describe the behaviour of these notions for a deep convolutional network in section~6, for Highway and smaller networks in section~7 and discuss analogies with an analysis of linear regions in input space of~\cite{TrianglesA} and \cite{TrianglesB} in section~8. In section~9, we give an example showing that our observables can be used without modification for 2-photon calcium imaging data of the mouse visual cortex. We close the paper with a discussion and some technical appendices.

\section{Related work and further motivations}
\label{s.relatedwork}

The notion of complexity is widespread in both natural and computer science. Yet it is often linked with quite diverse intuitions, hence the same term is used in a non-interchangable manner.
In this section, we would like to give a brief review of related work, emphasizing the similarities and differences with our approach. In addition, we also describe some developments in Physics which were the main motivation for our work.

\subsection{Foundations of Machine Learning}

In the machine learning context there are fundamental ideas on the model complexities, most prominently the VC dimension and Rademacher complexity. These measure the richness of the class of functions that can be learned by the given algorithm, and in practice play more formal than applied role when dealing with real world problems \cite{shai14}, \cite{Bengioi17}.

More model structure oriented notions are based on circuit complexity \cite{Roychowdhury90}, \cite{Parberry04}, where the neural network is treated as e.g. weighted majority, logical or threshold circuit and its expressive power is analyzed by e.g. relating network depth to the number of Boolean functions realizable by the network, or its resource usage scaling with input size.
This is quite different from what we do, as we aim to quantify the complexity of \emph{a particular} realization of a trained neural network, which computes some fixed function determined by its parameters and topology. This aspect is crucial for our approach, as we intend to investigate how complexity varies with the learning task while keeping the underlying network architecture fixed. 

\subsection{Motivations from Tensor Networks and Holography}
\label{s.holography}

In recent years an entirely different perspective emerged with the advent of tensor networks application to machine learning  \cite{Novikov15}, \cite{Garipov16}. Tensor networks are a way of representing large matrix multiplication structures of neural networks and play crucial role in quantum physics. They seem to be a bridge connecting starkly different areas of AI, quantum mechanics and many more \cite{Orus:2018dya}. In their realm one considers arithmetic complexity \cite{Austrin17} of tensor networks, their expressive power \cite{Cohen15}, \cite{Cohen17}, and incorporates fundamental concepts of quantum physics, like entanglement, representing correlations at different scales \cite{Cohen17a}, \cite{Lewenstein19}, \cite{alex2016exponential}, to infer better neural networks design. Even more, one can train models defined directly in tensor network form \cite{Schwab16} with accuracy comparable to standard deep neural networks. 
Physics based notions of complexity were developed primarily in the continuous setting (see e.g. \cite{Campo18}) but with variants of circuit complexity resurfacing~\cite{Heller}. 
These last developments were fuelled by another view on complexity which emerged from outside computer science, through so-called \emph{holography}.

A particularly fascinating phenomenon in theoretical physics is the principle of holography~\cite{holography}, where physical systems are described using an additional emergent dimension, in which physical features are encoded. 
Distance scales in the system are associated with locations in the ``holographic'' extra dimension (see~\cite{Hashimoto} for an analogy with neural networks). Certain tensor network constructions mentioned earlier, like MERA, realize something analogous~\cite{Swingle}, \cite{Orus14}. 

Indeed, one can understand the standard MERA tensor networks as providing a hierarchical, more and more non-local scale dependent parametrization of a quantum mechanical wave function. Holography, on the other hand, provides a rewriting of a quantum field theory in terms of other nonlocal degrees of freedom, with the nonlocality increasing as one goes into the extra dimension, associated also with scale. 

An analogous way of thinking about deep convolutional neural networks is that they provide something similar but for the complex probability distributions of natural images. Namely, features at a given depth (i.e. the output of a given layer in the interior of the network) provide a nontrivial nonlocal representation of the input images, with the nonlocality increasing with depth and culminating in the overall semantic classification at the output of the network.     

The above considerations motivate us also to investigate \emph{depth-wise} profiles of observables which probe the neural network at various scales and, moreover, to interpret the results not only in terms of properties of the given neural network, but also as nontrivial signatures of the structure of the particular dataset/classification task (analogous to a physical system) as a function of spatial extent scale and a hierarchy in nonlocality. 

In the holographic context, complexity is realized, according to one proposal, as the (spatial) volume of space in the above mentioned extra dimension~\cite{SusskindVolume} and thus as something additive over this extra holographic direction. Such a picture would suggest that a corresponding quantity in the neural network context would be additive over layers, if we interpret network depth as an analog of the holographic dimension "normal" to the directions along which input data extend (i.e. the image width and height).

The notion of complexity for neural networks that we propose in the current paper is not a direct translation of the above concepts but it draws on these intuitions like considerations on gate/operation cost and an additive layer-wise structure.

\subsection{Interrelations with Statistical Physics and Neuroscience}

A different line of motivation rooted in more conventional physical intuitions is offered by the following. The entropic character of our definition is grounded in physical considerations, especially in entropic properties of any computational process, and originally inspired by the Landauer principle  stating, that any irreversible computation entails entropy production \cite{Landauer}. 

In addition, the effective binary variables introduced in the present paper open up the possibility to use the methods of analysis developed for ensembles of spiking biological neurons~\cite{IsingNeurons}, modelled e.g. using a mapping to generalized Ising models~\cite{InverseIsing}, where the binary variables play the role of spins.
Along these lines, one could use similar methods of analysis based on statistical physics also to the case of deep neural networks. We leave this for future investigation.

We should note, that of course there exist already various studies which link together statistical physics and machine learning, among them the classical work of~\cite{Watkin}, which in fact discusses the complexity of a boolean rule as related to the number of boolean network realizations of this rule. This is, however, quite different from our approach. 

More recently one could mention e.g. the work~\cite{SGDNeurIPS}, where it is noted in particular, that generalization is linked to the interplay between optimizer, network structure and the dataset. This theme is aligned with the line of our approach, which aims at characterizing network \textit{and} dataset complexity (or difficulty) together. 


\subsection{Relation to selected contemporary Machine Learning approaches}

A prominent idea employing information theory notions in direct proximity to our entropic perspective is the Information Bottleneck (IB) approach to the understanding of deep learning \cite{infobottle}, \cite{saxe}. IB aims at quantifying network learning, optimal representation complexity and generalization properties (among many other things) based on both the dataset joint probability distribution $p(X, Y)$ and internal network representations $p(H)$ using mutual information $I(H,X)$ and $I(H,Y)$ between these quantities.

A common ground between our approach and the IB analysis is the emphasis on a joint treatment of the dataset/learning task with the hidden network representations. 
In contrast to IB, our approach however is based on entropy and operationally employs only the statistics of internal network representations, thus avoiding dealing with the computationally formidable task of estimating mutual information between the dataset and the hidden network representation\footnote{Also by using either binary variables (for \emph{complexity}) or continuous ones (for \emph{effective dimension}), we avoid the binning subtleties pointed out in~\cite{saxe}. They might reappear for non-ReLU activations, but we do not investigate that case in the present paper.}. We should emphasize, that the information about the dataset is still present in our approach, as the neural network representation statistics are \emph{induced} from the input dataset.
It would be certainly very interesting, however,  to investigate further the interrelations with the IB principle, but we leave this for future study.

A particularly interesting connection of our new notion of complexity can be made to the work quantifying neural network complexity by counting the number of linear regions of the function the neural network computes, \cite{TrianglesA} and \cite{TrianglesB}. There, the flexibility of the class of functions a given rectifier deep feed forward neural network can \textit{a priori} exhibit is related to the number of regions in the input data space, in which the network acts linearly. Such notion is readily associated with the piecewise liner approximation the network furnishes, and its exponentially growing accuracy (or flexibility) with the network depth \cite{TrianglesA}. The connection to this work is made by observing, that each linear region in the input space corresponds to a particular pattern of binary activations (which we define in the present work) in the network, and this pattern is in turn precisely captured by our analysis outlined below, but from a different perspective. We shall return to the signalled connection in Section~\ref{seq:triangles} and make it more explicit after having introduced our approach.

\section{Comments on convolutional neural networks}

\label{s.conv}

Before we describe the proposed observables, we should emphasize the way in which we will treat convolutional layers of a neural network. The individual computational unit of a convolutional layer is a set of $C$ neurons/filters (say of size $3\times 3$). The output of a layer for a batch of $N$ images has dimension $N \times C \times H \times W$. This output arises by passing through the convolutional filters a large number of small $3 \times 3$ patches of the output of the previous layer. Thus the neurons in a convolutional layer effectively see a batch of separate $N\cdot H\cdot W$ small patches of dimension $3\times 3$. Since we want to concentrate on analyzing the elementary independent computational units of the convolutional layer, in the main paper
we will always treat the output of the convolutional filters as $C$-dimensional, encoded in a huge effective batch of dimension
\eq
(N \cdot H \cdot W) \times C
\eqx
The only exception is a study in~\ref{sec.hwnet_app}, where we focus on particular pixels in the convolutional layer output. There, the output is still $C$-dimensional, but comes from an effective batch of dimension $N \times C$.


\section{Complexity -- entropy of nonlinearity and related notions}
\label{s.complexity}

The general idea of circuit complexity is to decompose a computation in terms of certain
elementary operations (\emph{gates}) and count the minimal number of gates (possibly with an appropriate cost
associated to a given gate) necessary to achieve that particular computation up to a given accuracy threshold.
We will not implement this prescription for neural networks but rather it will serve us as a heuristic motivation.

One clear insight from that approach is that it would not make too much sense to use a gate which, when composed multiple times could
be recast just as a single application of such a gate i.e.
\eq
A_1 \circ A_2 \circ \ldots \circ A_n = A 
\eqx
An example of such an operation would be a linear map, as a composition of many linear maps can be substituted by a single one. Hence from \emph{this point of view} it is natural to assume that a linear mapping is trivial and should not contribute to the estimation of complexity.

The most commonly used neuron in current artificial neural networks has a ReLU activation 
\eq
\label{e.relu}
y = ReLU(W_i x_i + b)
\eqx
Depending on the weights and the inputs, the neuron acts either as a linear function
$W_i x_i + b$ when $W_i x_i + b\geq 0$ or as a zero. In fact both options are trivial under composition.
Hence, the complexity of the neuron (\ref{e.relu}) 
should be associated with its decision function i.e. it acting
sometimes in the linear regime and sometimes clamping the output to zero.
To this end, let us introduce the binary variable
\begin{equation}
z = \theta(W_i x_i + b)
\end{equation}
where $\theta$ is the Heaviside function, so $z=+1$ if the neuron operates in the linear regime (i.e. it is ``turned on'') and $z=0$ if the output is set to zero and the neuron is ``turned off''. 
We will call this binary variable \emph{a~nonlinearity variable}.
Given the above intuitions, a very natural proposal for the complexity of a single neuron would be the Shannon entropy (measured in bits) of the nonlinearity variable $z$ computed on the dataset 
\eq
H(Z) = -\sum_{z=0,1} p(z) \log_2 p(z)
\eqx
where $p(z)$ is the probability distribution of the binary variable $Z$ obtained by passing examples from the given learning task/dataset through the neural network which contains the given neuron.
The above definition vanishes both for the purely linear and purely turned off regimes, and is sensitive to the switching behaviour of the neuron, i.e. its decision function.

We should emphasize that this notion of complexity of the ReLU neuron is dependent on the given dataset or learning task and on its place within the whole neural network. It measures the nonlinearity of the neuron's operation when it functions as a part of a neural network trained on that particular dataset and processing examples from that dataset.
Clearly, for a single neuron this measure is really quite trivial, but it becomes much more interesting when we consider a layer of neurons as a whole.

For other activation functions e.g. for a \emph{sigmoid}, a natural extension would be to define an auxiliary \emph{ternary} variable, corresponding to the three distinct regimes of operation: saturation at 0, approximately linear operation and saturation at 1, and use again the definition in terms of Shannon entropy. We will not, however, explore such generalizations in the present paper.

In order to define complexity for a set of neurons (e.g. forming a layer), we define a corresponding set of binary activations
\eq
z_k = \theta(W_{ki} x_i + b_k)
\eqx
and define the complexity of the layer as the entropy of the \emph{multidimensional} binary nonlinearity variable
\eq
\label{e.HZ}
complexity = H(Z_L)  = -\sum_{ \{\text{All}\ z_k=0, 1\} } p(z_1, \ldots, z_C) \log p(z_1, \ldots, z_C)
\eqx
where we collectively denote the considered set of neurons (usually forming a layer $L$ in a neural network) by $Z_L$. 
This definition captures also the important information whether the individual neurons switch their mode of operation
independently or not. 

Let us also note that such a measure is used as an indicator of the information content
of ensembles of spiking neurons (which by themselves are naturally binary) in neuroscience (see section~\ref{s.relatedwork}). 
A related analogy of the binary variables is with Ising spin variables.
This leads to new possible neuroscience/physics
inspired modes of analysis of artificial neural networks.

Our definition of complexity is thus essentially the entropy of nonlinearity of the neural network or layer:
\begin{center}
{\bf {Complexity} $\equiv$ entropy of {nonlinearity}}
\end{center}

In general, the evaluation of (\ref{e.HZ}) is not trivial. In practice, we cannot use
entropy estimation based on occurrence counts due to the huge dimensionality $2^C$ of the space of configurations.
In this paper we use, therefore, the method of estimation of entropy from~\cite{mlentropy} using the {\tt xgboost} classifier~\cite{xgboost} (see \ref{s.entropy} for more details).

Moreover, one should be careful with properly interpreting formula (\ref{e.HZ}) in the case of convolutional
neural networks. As described in section~\ref{s.conv}, $C$~is then just the number of channels and \emph{not} the 
total number of outputs $C\times H \times W$. This makes it feasible to evaluate the entropy for all layers of a deep convolutional network such as resnet-56, which we study as an example.

For the whole network, one could in principle use a similar definition
\eq
\label{e.complexitytotal}
complexity(network) = H\left( \bigcup_{L \in Layers} Z_L \right)
\eqx
but that would be quite difficult to
reliably evaluate for deep networks due to the dramatically larger dimensionality of the overall configuration space. For convolutional networks there would be a further complication due to the different resolutions in various parts of the network and hence different numbers of patches for different groups of layers. So the definition (\ref{e.complexitytotal}) is directly applicable only for nonconvolutional networks.

However, as emphasized in the introduction and in section~\ref{s.relatedwork}, it is even more interesting to study the variation of the complexity of individual neural network layers as a function of depth. This information describes the nonlinearity of processing as a function of \emph{scale} and can be understood in a dual way as a characteristic of a given learning task/dataset.
In the present paper, we will therefore mostly concentrate on the individual layer complexities $H(Z_L)$ (or their averages over blocks of layers) as a function of depth, training epoch and dataset.
Continuing along these lines, for the whole network we may \emph{define} the \emph{additive complexity} of a network by adding up the complexities/entropies of individual layers:
\eq
additive\ complexity(network) = \sum_{L \in Layers} H(Z_L)
\eqx

\subsection{Derived quantities}

It is useful to consider also some other quantities derived from the same data. Although Shannon's entropy of a set of binary variables is clearly sensitive to their interdependence structure, it is convenient to use an observable which \emph{explicitly} measures that interdependence. 
One natural candidate is the \emph{total correlation} (or multiinformation)
between the neurons in a given layer 
which captures the extent in which the individual neurons switch independently from each other. We will use it in the normalized variant
\eq
\label{e.totcorrnorm}
normalized\; total\; correlation = \f{H(Z_1)+\ldots + H(Z_C) - H(Z_1, \ldots Z_C)}{H(Z_1)+\ldots + H(Z_C)}
\eqx
Another interesting variable is the average \emph{linearity} of a layer defined by
\eq
linearity = \f{1}{C} \sum_{i=1}^C E(Z_i)
\label{e.linearity}
\eqx
where $E(Z_i)$ is the expectation value of the nonlinearity variable (evaluated on the test set of the corresponding dataset).
The advantage of this quantity is that it is very efficient to compute (at the cost of ignoring correlations) and it distinguishes predominantly linear operation 
of a neuron from predominantly being turned off, two conditions which are on the same footing and thus indistinguishable from the point of view of entropy and hence complexity.
The above suggests, that one should also quantify the properties of the \emph{linear} operation of a neural network layer. 
In the following section, we will introduce an observable complementary to complexity which aims to do just that.

\section{Effective dimension}     
\label{s.dimeff}

The particular definition of complexity introduced in section~\ref{s.complexity} measures the inherent \emph{nonlinearity} of the operation of the neural network.
It thus ignores any properties associated with the linear mode of operation of the network. In particular, if all the neurons in a layer would operate in the linear regime, the corresponding complexity would exactly vanish. Yet, the pattern of their activations would describe potentially nontrivial feature representations. Hence, we would like to have a complementary observable, which would be sensitive to that. 
To this end we introduce the notion of an \emph{effective dimension} of the feature representation of the given layer.

Suppose that a layer has $C$ output channels. We perform PCA of the $C$-dimensional layer output (understood as in section~\ref{s.conv} in the convolutional case) obtaining the principle components and their appropriate variance ratio explained 
\eq
r_i=\f{\lm_i}{\sum_{i=1}^C \lm_i} \qq \text{with} \qq \sum_{i=1}^C r_i = 1
\eqx
The conventional method of estimating dimensionality would be to set some threshold for the total variance ratio explained and count the number of components which suffice to explain this threshold\footnote{This was the route taken in~\cite{pcann} with an explicit threshold for the explained variance ratio. A more significant difference with the observable defined in that paper is that we consider the output of each patch individually (see section~\ref{s.conv}) while~\cite{pcann} used average pooling over the whole image.}. The drawback is that this procedure introduces an arbitrary auxiliary parameter. Bayesian methods provide an alternative, more principled way, but are much more involved~\cite{PPCA}.
Here we would like to introduce a new, very simple parameter-less estimation of the effective dimensionality of the output. 

To this end, we {\it formally} interpret $r_i$ as a probability and evaluate the Shannon entropy\footnote{Note that here we use the natural logarithm and not $\log_2$ as previously.} of $r_i$:
\eq
S[\{r_i\}] = -\sum_{i=0}^C r_i \log r_i
\eqx
Subsequently, we define the effective dimension as
\eq
\label{e.effdim}
ef\!fective\; dimension = \exp \left( S[\{r_i\}] \right)
\eqx
This definition reproduces the correct answer ($ef\!fective\; dimension=k$) for simple cases like 
\eq
r_1=r_2=\ldots=r_k=1/k, \qq r_{i>k}=0 \qq \text{with} \qq k \leq C
\eqx
which would correspond to a $k$-dimensional ball embedded in the $C$-dimensional space of activations. The definition (\ref{e.effdim}) provides a natural interpolation from these simple cases to the generic situation.
Also it mimics the physical picture of entropy as the logarithm of the number of microstates.


\section{Complexity and effective dimension for a deep convolutional neural network}


In this section we perform experiments with a resnet-56 network (version for CIFAR-10 \cite{resnetcifar}). The network is structured as follows: an initial convolutional layer (block 1), then three blocks of 9 residual cells each, and an output layer following average pooling. 
The three blocks (blocks 2-4) operate on resolutions $32 \times 32$, $16\times 16$ and $8\times 8$ respectively and have 16, 32 and 64 channels. 
The variant of the residual cell used here has a ReLU right after summation. The second ReLU is in the residual branch. For the observables defined in sections \ref{s.complexity} and \ref{s.dimeff} we can collect the outputs after either ReLU. In the present paper we discuss the outputs of the ReLU after summation (to which we refer as the `topmost' one). 
This can be understood as measuring the \emph{complexity} and \emph{effective dimension} of the whole residual cell treated as a single unit as it is exactly this output which goes into subsequent residual cells.

The outputs are always evaluated on the test sets of the respective datasets, apart from CIFAR-10 with randomized labels, where we use a sample of 10000 images from the training data.
The training protocol is the same for all datasets\footnote{The only exception is that weight decay is set to zero for CIFAR-10 with randomized labels.} (see~\ref{s.training} for more details).
For all datasets we perform 5 simulations with different random initializations. For the case of memorization i.e. CIFAR10 with randomized labels, the network achieved good training accuracy only for 3 initializations (reaching accuracies 93.68, 98.40, 99.10), while for the 2 other initializations it failed to learn in the allotted 100 epochs (reaching accuracies 24.17 and 33.62). In the following we separately analyze the two cases.

\subsection{Properties of trained networks}

\begin{figure}[t]
  \centering
\hfill\includegraphics[height=5cm]{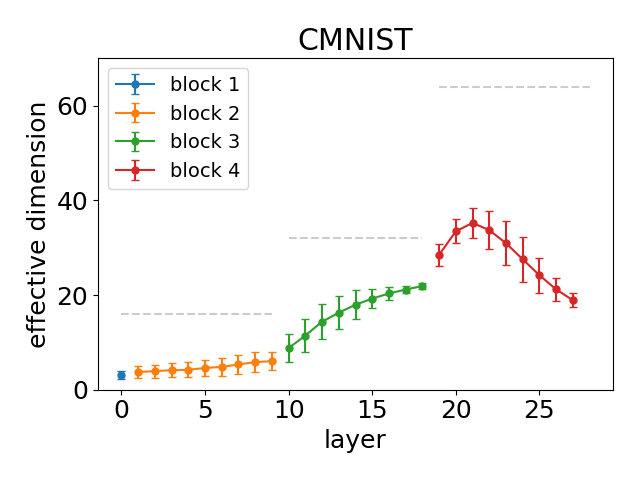} \hfill
\includegraphics[height=5cm]{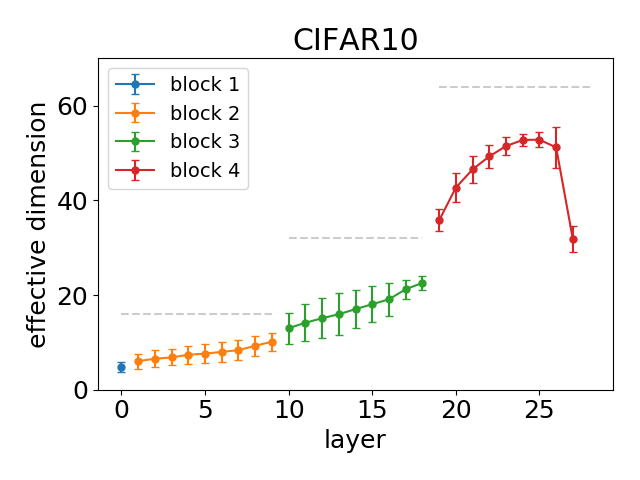} 
\hfill\mbox{}\\
\hfill\includegraphics[height=5cm]{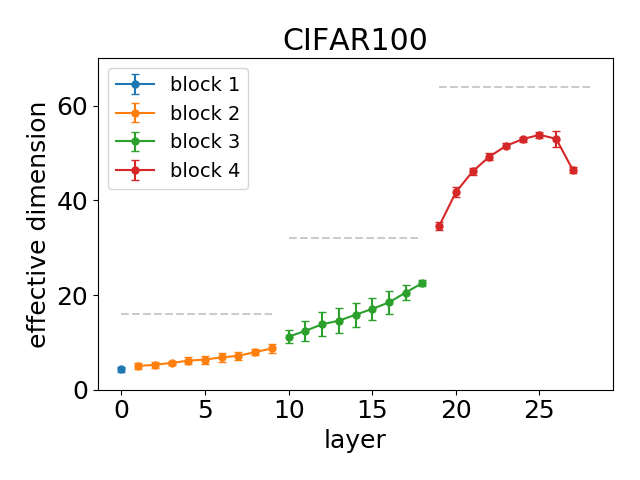} \hfill
\includegraphics[height=5cm]{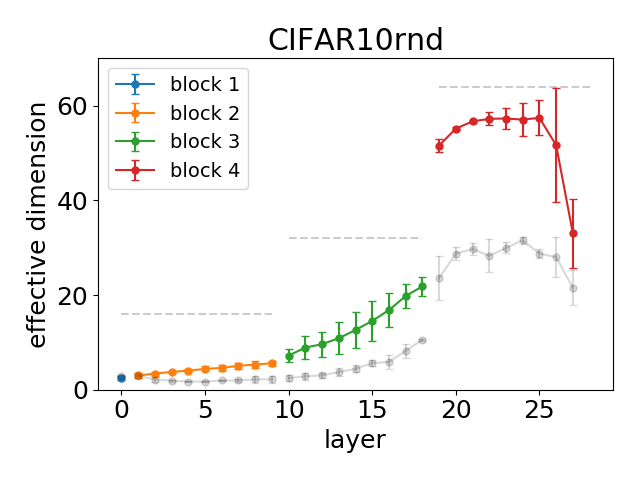} 
\hfill\mbox{}
\caption{``Holographic'' plot of the \emph{effective dimension} as a function of layer depth (topmost ReLUs in residual cells) numbered from the input. The distinct blocks with different number of neurons are marked with different colours. The dashed lines represent the number of neurons in each layer. For CIFAR10 with randomized labels we show separately the results for the successful networks and the failed ones (shown in gray).}
\label{fig.effdimholo}
\end{figure}

\begin{figure}[t]
  \centering
\hfill\includegraphics[height=5cm]{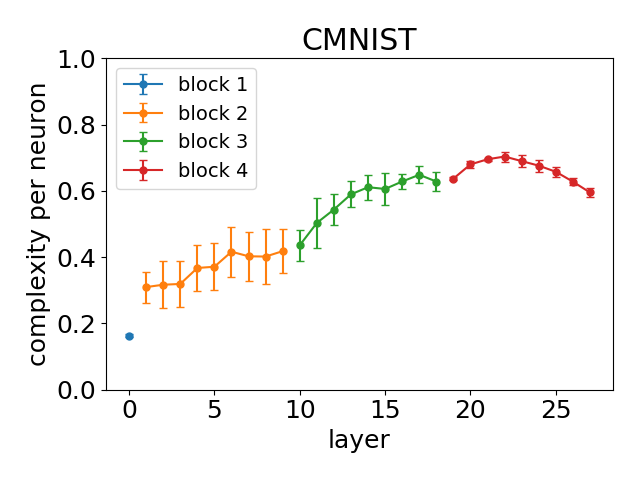} \hfill
\includegraphics[height=5cm]{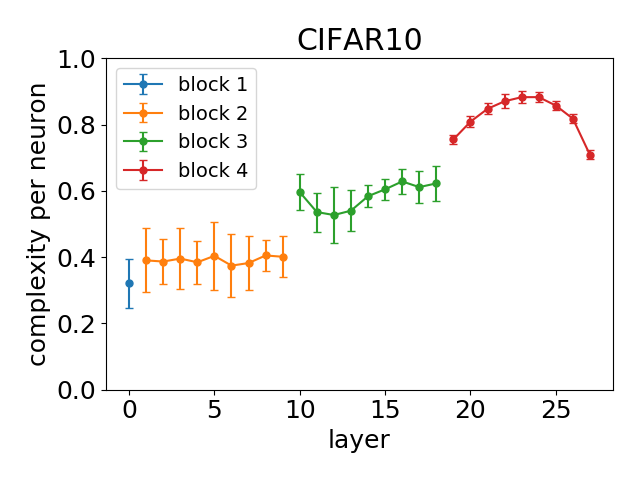} 
\hfill\mbox{}\\
\hfill\includegraphics[height=5cm]{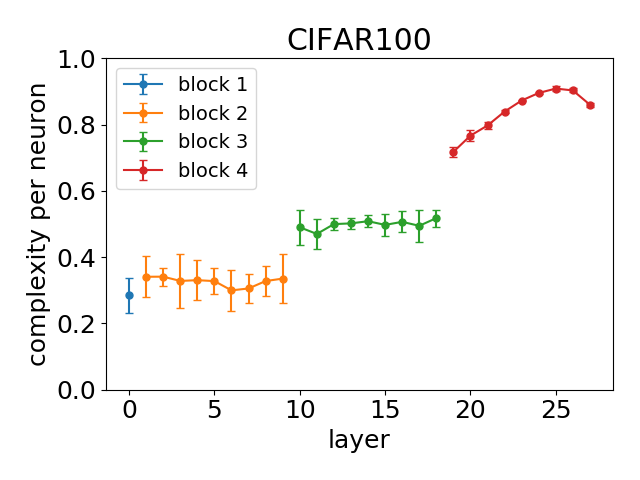} \hfill
\includegraphics[height=5cm]{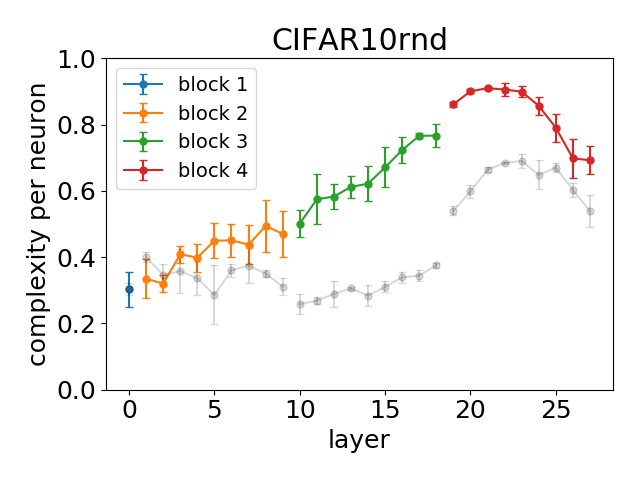} 
\hfill\mbox{}
\caption{``Holographic'' plot of the \emph{complexity per neuron} as a function of layer depth (topmost ReLUs in residual cells). For CIFAR10 with randomized labels we show separately the results for the successful networks and the failed ones (shown in gray).}
\label{fig.complexityholo}
\end{figure}

\paragraph{Depth-wise hierarchy} In Figure~\ref{fig.effdimholo} we show the dependence of the \emph{effective dimension} as a function of layer depth for CMNIST, CIFAR-10, CIFAR-100 and CIFAR-10 with randomized labels. For the conventional datasets, one can clearly see  a consistent relatively smooth profile of increasing dimensionality of the feature representations, even though the three distinct stages with different resolution have markedly different number of channels (16, 32 and 64 respectively). 

In the case of memorization (Fig.~\ref{fig.effdimholo} lower right), we see a distinct pattern appearing, with high effective dimension in the topmost block, while remaining quite low in the lower resolution stages of processing. Note that the networks which did not manage to learn have significantly smaller effective dimensions, especially in the topmost block.

Overall, we observe a significant rise in the dimensionality of the highest level feature representations with the difficulty of the task.

Figure~\ref{fig.complexityholo} shows the corresponding plot with \emph{complexities} normalized to the number of channels. Here 
we observe a much flatter behaviour on CMNIST
and an increase of the complexity per neuron as we go to higher stages of processing for the CIFAR-10 dataset. Note that in contrast to the \emph{effective dimension}, \emph{complexity} scales more naturally with the number of channels/neurons. 
This can be most clearly seen in the plot for CMNIST (Fig.~\ref{fig.complexityholo} (upper left)), where the complexity per neuron is virtually continuous across block boundaries, where the number of neurons increases by a factor of 2.

For the case of memorization (Fig.~\ref{fig.complexityholo} lower right), we see generally relatively high levels of \emph{complexity} throughout the network, which is in marked contrast with the behaviour of  \emph{effective dimension}.
But again we can see a very significant difference between the networks which successfully memorized the training set and those that didn't.


To summarize the above discussion,
we observe that complexity and effective dimension depth-wise hierarchy of the datasets, as seen by a resnet-56 network, exhibit an increasing 
tendency to represent abstraction in data while going to larger spatial scales as manifested by higher nonlinearity (on/off switching) and richer (higher dimensional) feature representations. The eventual clear decrease towards the final layers is most probably due to the fact that the output\footnote{Note that in all plots we show only the convolutional layers prior to average pooling and the output layer.} of the network has 10 classes and the features have to be integrated to arrive at the final classification answer. Note that the corresponding dip for CIFAR-100 in Fig.\ref{fig.complexityholo} is much smaller.

Finally, let us emphasize that one can interpret all the above observations from two points of view. On the one hand, one can gather insight into the finer details of how the deep neural network deals with the particular classification task. On the other hand, one can view these results as representing nontrivial information about the internal structure of each particular dataset at various scales.
Thus one could view the plots in Figs.~\ref{fig.effdimholo}-\ref{fig.complexityholo} as representing a Deep Neural Network based transform or characteristic of the whole learning task/dataset. We called the plots ``holographic'' in order to underline the analogy touched on in section~\ref{s.holography}.

\begin{figure}[t]
  \centering
\hfill\includegraphics[width=0.6\textwidth]{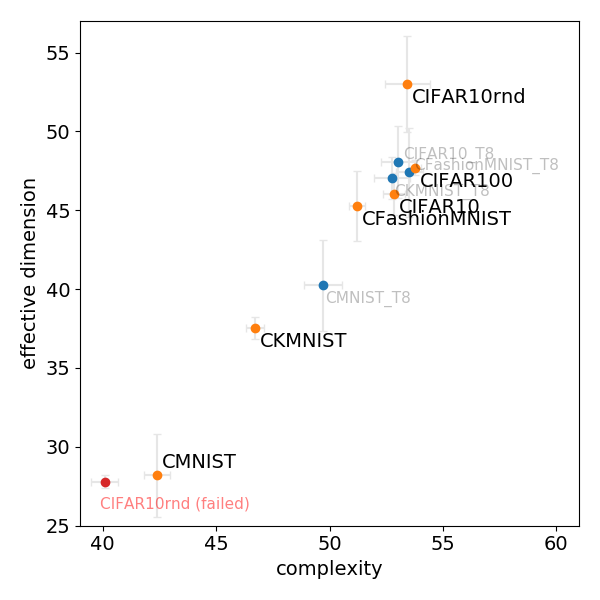}\hfill\mbox{}
\caption{\emph{Effective dimension} vs. \emph{complexity} (averaged over the top block) for trained resnet-56 on various datasets after 100 epochs. For CIFAR10 with randomized labels we separately consider the successful and unsuccessful networks.}
\label{fig.effdimcomplex}
\end{figure}

\begin{figure}[t]
  \centering
\hfill\includegraphics[width=0.47\textwidth]{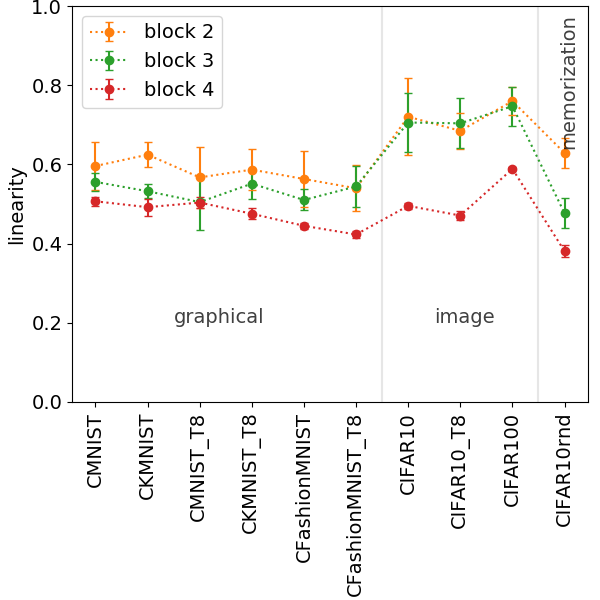} \hfill
\includegraphics[width=0.47\textwidth]{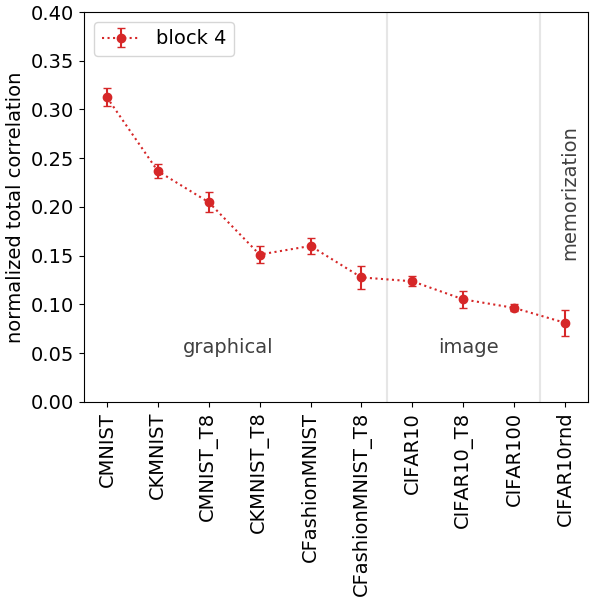} \hfill\mbox{}
\caption{Average \emph{linearity} for layers in the three successive blocks of resnet-56 for various datasets (left). Normalized total correlation (right). The lines serve to guide the eye. For CIFAR10 with randomized labels, this includes only the successful 3 networks.}
\label{fig.linearitytotcorr}
\end{figure}

\paragraph{Dataset difficulty} In order to see how the \emph{effective dimensions} and \emph{complexities} are correlated with the difficulty of the learning task, we plot these observables in Fig.~\ref{fig.effdimcomplex}  for a selection of datasets. These observables (for the topmost ReLU) are  averaged over the layers in the top block of the network, which encompasses the high level processing.

We observe a clear correlation with the intuitive difficulty of the classical datasets. The teared up versions are also more to the top and right than the original ones as one would intuitively expect.
The successful and unsuccessful memorization task networks situate themselves on the opposite extremes of the plot. In this respect, we emphasize that neither \emph{effective dimensions} nor \emph{complexities} involve any knowledge about the target labels.
An additional intriguing feature of this plot is the fact that the various datasets lie on an approximately linear band in the \emph{effective dimension}-\emph{complexity} plane. 

It is interesting to understand in more detail the nature of the rise of complexity with the difficulty of the dataset. Such a rise could \emph{a-priori} arise in two scenarios: i) the probability of switching of individual neurons could approach $p=0.5$, which maximizes the \emph{individual} neuron entropy/complexity or ii) the increase of complexity arises mostly due to the switching of the neurons becoming more independent.

In order to study this problem,
in Fig.~\ref{fig.linearitytotcorr} (left) we plot the averaged \emph{linearity} in the three successive blocks of layers of resnet-56 for the various datasets. As a side remark, we see that for the image datasets the lower blocks operate at a higher linearity, while for the memorization task they revert back to the lower values. For the question at hand, we observe that at the highest level of processing (block 4) the mean linearity is approximately the same for all datasets, being close to the optimal value $0.5$ already for the simplest datasets. 
This shows that the rise of complexity does not follow from scenario i).

Indeed, we observe a clear decrease of the normalized total correlation (\ref{e.totcorrnorm})
with the difficulty of the dataset as shown in Fig.~\ref{fig.linearitytotcorr} (right). This confirms scenario ii) and indicates that the higher level
neurons operate in a much less correlated way for the more complicated datasets and so perform more independent decisions.

\begin{figure}
  \centering
\hfill
\includegraphics[width=0.47\textwidth]{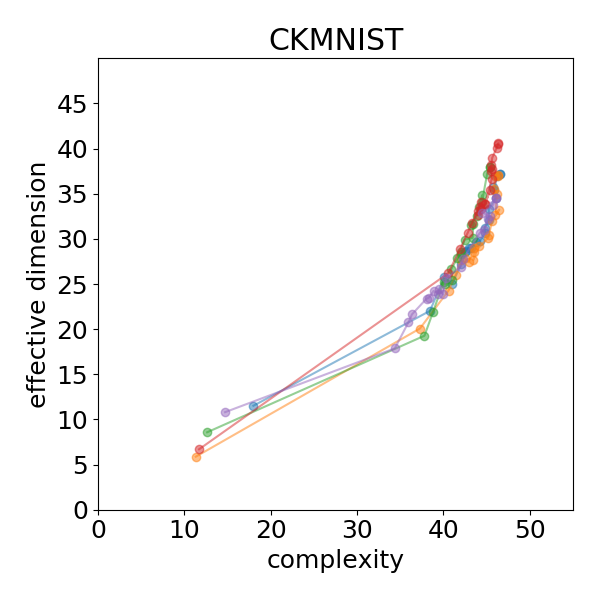} \hfill\mbox{}
\includegraphics[width=0.47\textwidth]{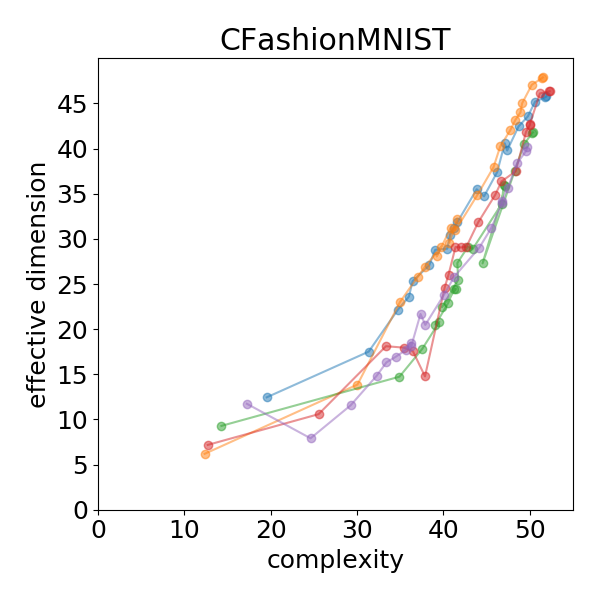} \hfill\mbox{}\\
\hfill
\includegraphics[width=0.47\textwidth]{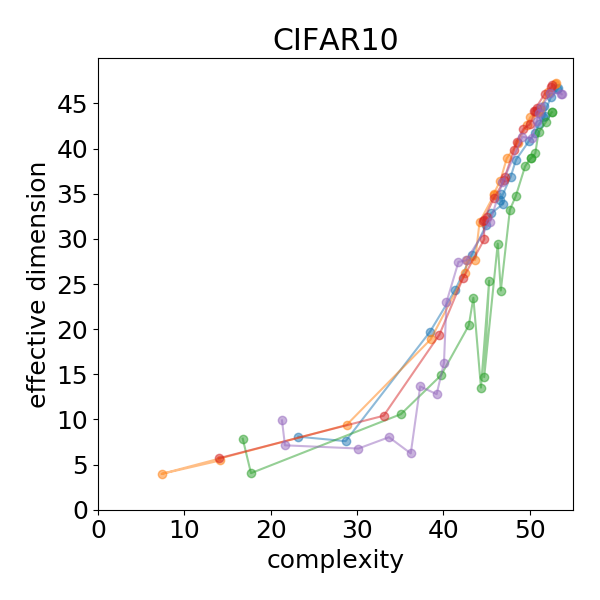} \hfill\mbox{}
\includegraphics[width=0.47\textwidth]{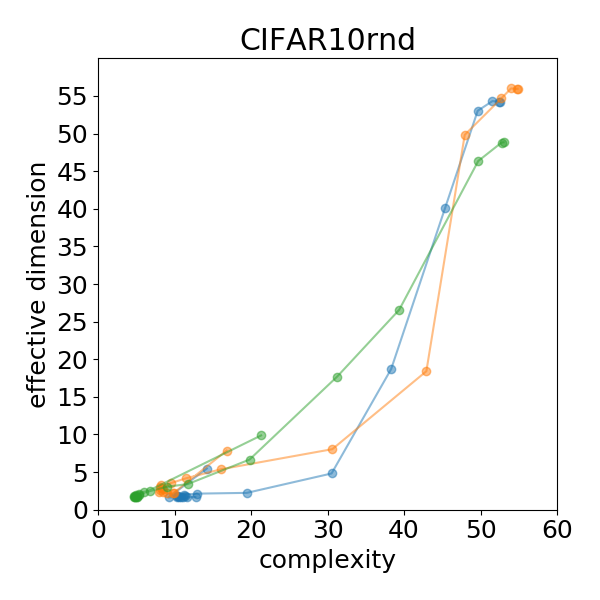} \hfill\mbox{}
\caption{Training trajectories in the \emph{effective dimension} - \emph{complexity} plane for 5 different random initializations for CKMNIST, CFashionMNIST, CIFAR10 and CIFAR10 with randomized labels (in this case only for 3 out of 5 initializations the network managed to learn in the allotted time). The dots denote epochs 0 (initialization before training), 1-10, 15, 20, 30, \ldots, 100. The observables are averaged over the layers in the topmost block.}
\label{fig.trajectories}
\end{figure}

\subsection{The dynamics of training}

\begin{figure}[t]
  \centering
\hfill\includegraphics[width=0.47\textwidth]{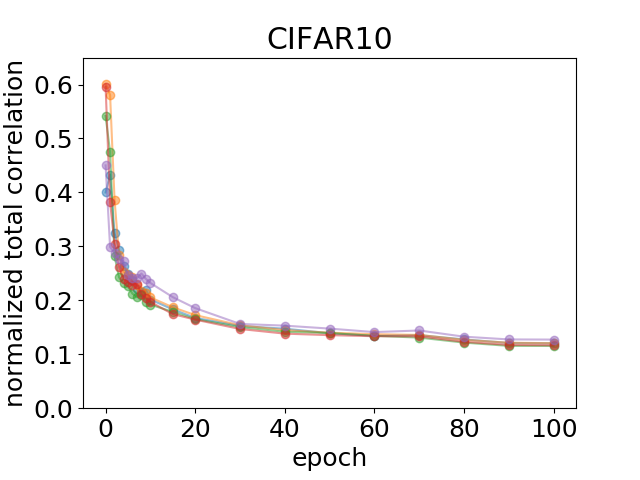} \hfill
\includegraphics[width=0.47\textwidth]{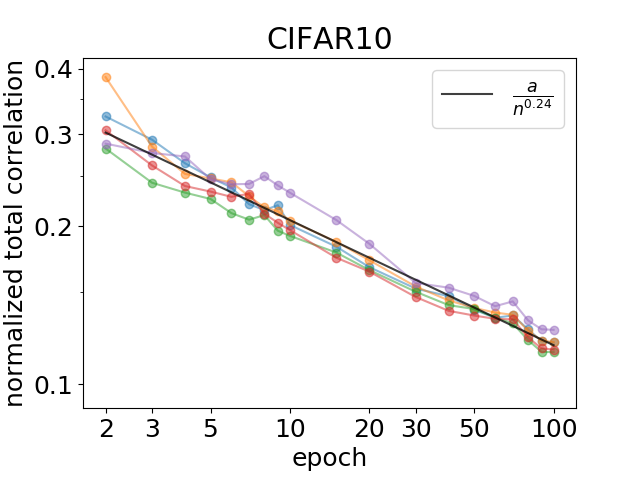} \hfill\mbox{}
\caption{Evolution of normalized total correlation during training for 5 different random initializations (left). Log-log plot with a power law fit (right).}
\label{fig.totcorrfit}
\end{figure}

As indicated in the introduction, it is very interesting to analyze the evolution of the introduced observables during the learning process. In Fig.~\ref{fig.trajectories} we show the trajectories in the \emph{effective dimension}-\emph{complexity} plane for the topmost layers (block 4) for a selection of datasets and different random initializations.

We observe two distinct stages of training. Initially the \emph{complexity} tends to increase fastest with only a moderate rise in the \emph{effective dimension}. Then both observables increase, but now the \emph{effective dimension} increases in a much more significant way. The length of the initial stage is relatively short and varies with the dataset \emph{and} initialization. An exception is the case of memorization, where this initial stage is quite extended and the transition occurs only around 30 epochs.
Overall, we observe that during learning the network generates richer (effectively higher dimensional) high level representations and builds up nonlinearity. 

Let us note that the picture of training dynamics shown in Fig.~\ref{fig.trajectories} involves observables which are defined purely locally i.e. within individual layers, and thus is complementary to the style of analysis in the framework of Information Bottleneck~\cite{infobottle}, which involves the mutual information of a layer feature representation either with the network input or with the output target class.

\begin{table}[]
    \centering
    \begin{tabular}{l|c}
dataset   &  $\alpha$ \\
    \hline
CKMNIST   & 0.07 \\
CFashionMNIST & 0.22 \\
CIFAR10 (slow) & 0.16 \\
CIFAR10 & 0.24 \\
CIFAR10rnd & 1.52
    \end{tabular}
    \caption{The exponent $\alpha$ characterizing the power law scaling of normalized total correlation during training. CIFAR10 (slow) indicates training with the learning rate decreased by a factor of 2. For CIFAR10rnd, the fit was performed only for epochs 40-80.}
    \label{tab.alpha}
\end{table}

As we have seen in the previous section, the increase of \emph{complexity} in the highest layers occurs through the decisions of the neurons being more and more independent as measured by the normalized total correlation (see Fig.~\ref{fig.linearitytotcorr}). Hence it is very interesting to study the behaviour of this quantity during training. Its evolution for CIFAR-10 is shown in Fig.~\ref{fig.totcorrfit}. We observe a very clear decrease during training, showing that the individual high level features become more and more independent. Moreover, the decrease follows a power law with the number of epochs $n$ (see Fig.~\ref{fig.totcorrfit} right):
\eq
normalized\ total\ correlation \sim \f{A}{n^\alpha}
\eqx
This conclusion holds also for other datasets. The values of $\alpha$ are shown in Table~\ref{tab.alpha}. The power $\alpha$ is not universal for the neural network but depends on the dataset and on the learning rate. 
However, the overall power law scaling for a particular dataset is very clear and captures an intriguing property of how the deep convolutional network is learning its highest level features. This poses an intriguing challenge for our understanding of deep learning.

\section{Fully Connected, Highway and small neural networks}
\label{s.shallow}
Clearly the observables defined in the present paper may be investigated for a network of any size and architecture. In particular, in this section we present complementary investigations focused on entropic features in minimal fully connected networks, in order to be able to explore variability with the network size and possible novel phenomena for other architectures.

\begin{figure}
  \centering
\hfill\includegraphics[height=4.52cm]{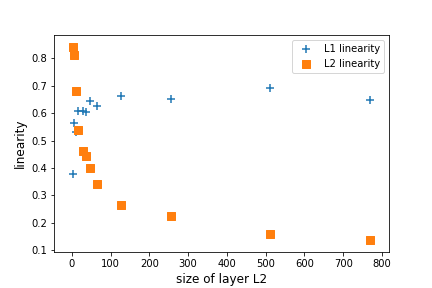} \hfill
\includegraphics[height=4.52cm]{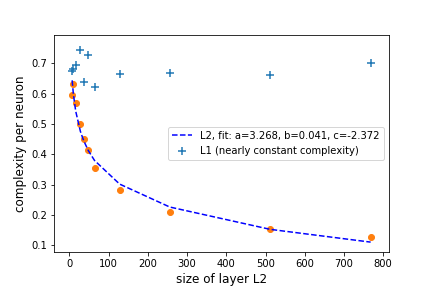} \hfill\mbox{}
\caption{Small FC3 network. Average \emph{linearity} scaling with $L2$ width $N$ (left). \emph{Complexity per neuron} scaling ($a/N^b+c$) of the second layer $L2$, with almost constant complexity of the first layer $L1$ (right).}
\label{fig.fc3entropy}
\end{figure}

\paragraph{Shallow fully connected network.} We consider a three-layer network (FC3)
with the entry layer $L1$ fixed at 14 neurons and the size of the middle layer $L2$ varying in the range $N\in[4, 768]$. All variants are trained on MNIST dataset to accuracy 95\%.
We find that the average \emph{linearity} changes substantially with $L2$ width for both layers as seen in the Fig. \ref{fig.fc3entropy} (left), even though the size of layer $L1$ is kept fixed. 
Interestingly, the \emph{complexity per neuron} for layer $L2$ scales as a power law, whereas for $L1$ it does not really depend on the size of the subsequent layer (Fig. \ref{fig.fc3entropy} (right)).

\begin{figure}
\begin{multicols}{2}
\centering
\includegraphics[width=1\linewidth]{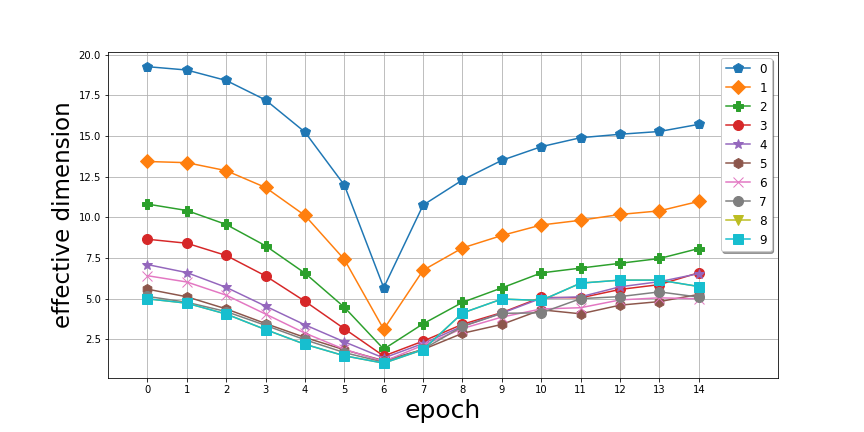}\\
\includegraphics[width=1\linewidth]{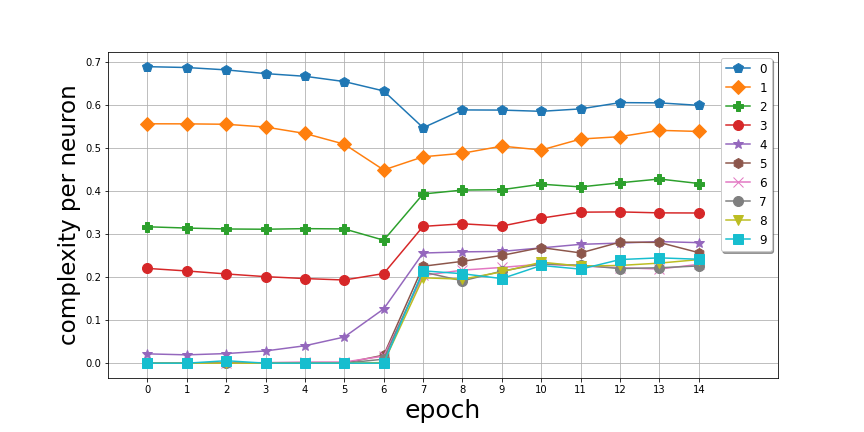}\\
\end{multicols}
\caption{11-layer 28 neuron wide Highway Network \emph{effective dimension} (\emph{left}) and complexity per neuron of the nonlinear part (\emph{right}). For the corresponding rapid loss transition see Fig. \ref{fig.PhaseTransition_App}. Notice also the ordered structure of profiles representing observables associated with respective network layers.}
\label{fig.PhaseTransition}
\end{figure}

\paragraph{Highway networks}
\label{seq:hwnets}
As an example of a deeper fully connected network with a nontrivial architecture we investigate an 11-layer, 28 neuron wide highway network~\cite{highway}. We observe a very distinctive pattern, reminiscent of a phase transition, in the behaviour of both the \emph{effective dimension} and network layers non-linear outputs \emph{complexity} during the course of training, Fig.~\ref{fig.PhaseTransition}, which coincides with the abrupt transition around epoch 7 in the test loss and accuracy (see Fig.~\ref{fig.PhaseTransition_App}). Interestingly the behaviour of the various layers gets slightly desynchronised and shifted relative to earlier and later epochs, once one changes the depth or width of the network (see Fig.~\ref{fig.three_hwnets}). Also noticeable is the preserved order of the structure reflecting network layers, in which corresponding profiles form the transition. This dip, or quench structure is not seen in the case of resnet-56, but can be observed in a small convolutional network at very low learning rate (see Fig.~\ref{fig.cnn_pixelwise}). 
Some additional comments on small convolutional networks are presented in \ref{sec.hwnet_app}.

\section{Relation to the linear regions analysis}
\label{seq:triangles}

Let us now briefly return to the relation between our work and the measure of complexity defined through the number of linear regions in input space, which we touched on in section~\ref{s.relatedwork}. As introduced in \cite{TrianglesA} one imagines, that the function represented by the whole rectifier neural network (with a specific set of weights) is performing hyperplane partition in the space of all possible input images. 
The partition follows from the consecutive compositions of the ReLU activations, defining linear regions as the neighbourhoods in the input space, in which the network acts linearly. Each input sample then falls into some segment of the partition and is associated with a particular pattern of activations of the network. This notion is very close to our approach of defining the binary indicator variable $z$ expressing non-linearity, whenever a given neuron fires or not. Therefore each linear region can be understood as corresponding to a particular activation pattern of our binary $z$ variable for all layers of the neural network.
Passing through the boundary to another linear region would correspond to flipping one of the $z$'s.

The measure of complexity introduced in \cite{TrianglesA}, \cite{TrianglesB} relies on counting the number of such linear regions, while our approach to complexity employs the entropy of the network non-linearity, but if the two are to be related in some qualitative way, one should expect some similarity between complexity estimations following from both notions. And indeed this is what we observe. 

\begin{figure}[t]
\centering
\hfill\includegraphics[height=5cm]{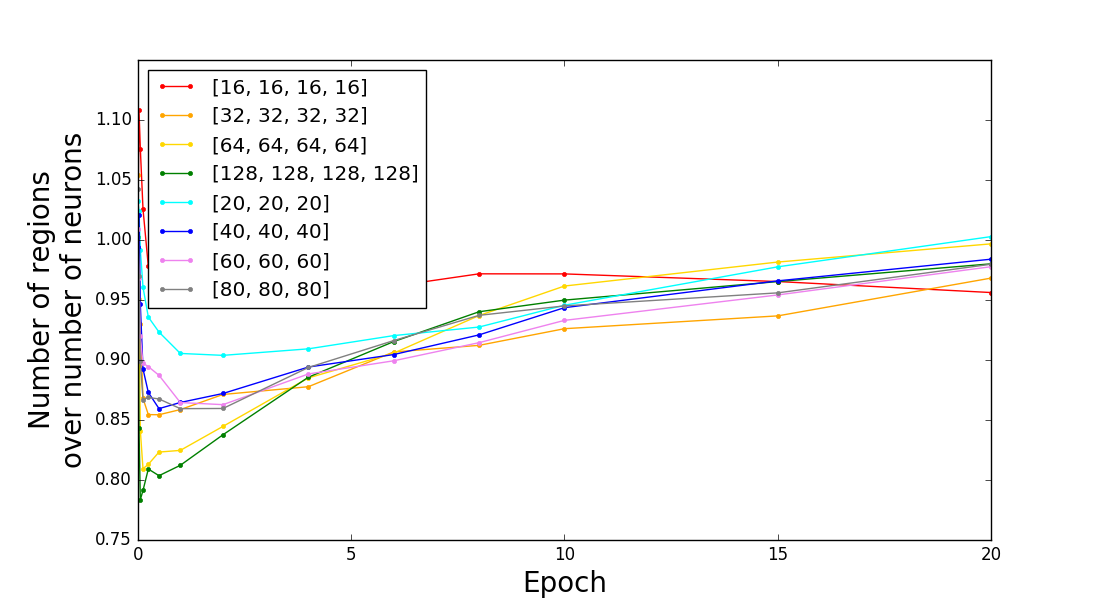} \hfill\mbox{}
\caption{Saturation of the linear regions count for various small FC networks widths classifying MNIST (taken from \cite{TrianglesB}).}
\label{fig.saturationTri}
\end{figure}

\begin{figure}[t]
\centering
\hfill \includegraphics[height=4.7cm]{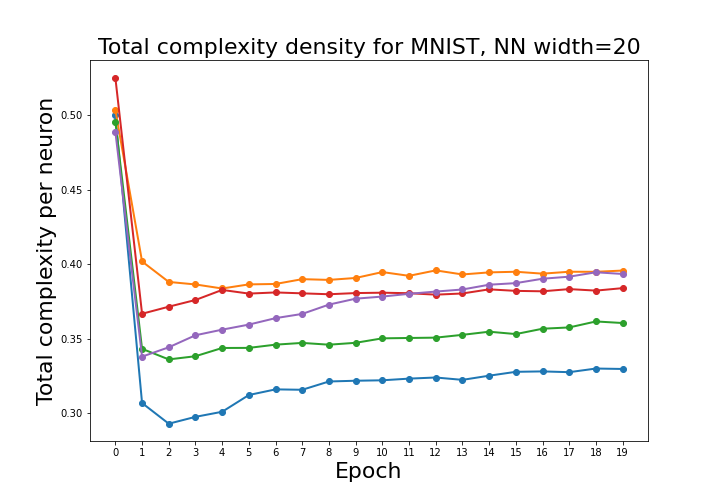} \hfill
\includegraphics[height=4.7cm]{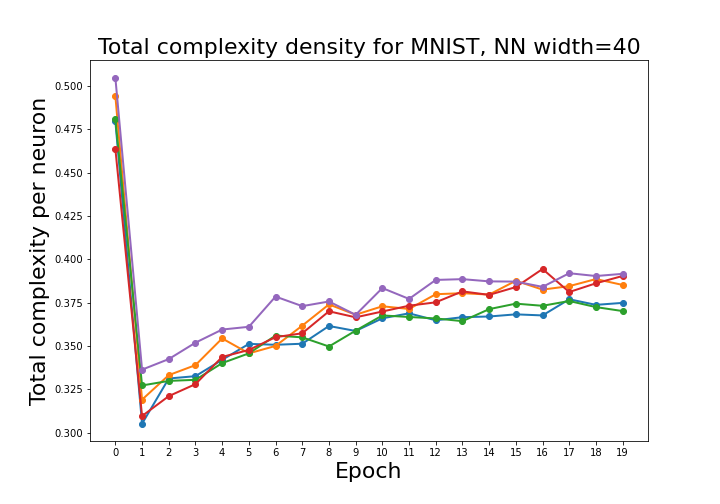} \hfill\mbox{}
\caption{Our observation of normalized complexity saturation for FC nets of widths 20 and 40 and several random seeds. The behaviour is qualitatively similar to the one for linear regions, Fig.~\ref{fig.saturationTri}, including approximate width independence: both lines settle around $0.37$.}
\label{fig.saturation}
\end{figure}

Figure \ref{fig.saturationTri} (from \cite{TrianglesB}) depicts the evolution of the number of linear regions during training for small fully connected neural networks of various widths. As the authors of~\cite{TrianglesB} point out, there is an apparent saturation in the observable, and to a degree independence from the network width (upon normalization). In Fig.~\ref{fig.saturation} we recreate exactly the setup  leading to Fig.~\ref{fig.saturationTri}, but show the behaviour of our notion of complexity. We notice the evolution of total normalized network complexity for widths 20 and 40 exhibiting analogous saturation and approximate width independence as for the linear regions approach. This observation also corroborates our interpretation of the $z$ entropy as model complexity. 

We would like to emphasize, however, that
there are certain operational benefits behind out notion of complexity. It is explicitly and easily calculable for any ReLU based architecture, with arbitrary topology, including very deep and convolutional networks. In fact out notion can be trivially generalized to any activation with countable number of irregularities (like the behaviour of ReLU at zero). Further, it is very granular as we can focus on arbitrary subsets of the network neurons, like particular layers, which provides information on the neural network processing at a particular scale or depth. 
Furthermore, one could readily restrict our observables to input data belonging to a particular class etc. 
The explicit calculability of our observables in contrast to theoretical bounds is especially important as we observe a marked dependence of complexity on the dataset for a neural network with a fixed architecture (resnet-56).

\section{Biological data -- the mouse visual cortex}

\begin{figure}[t]
  \centering
\hfill\includegraphics[height=4.48cm]{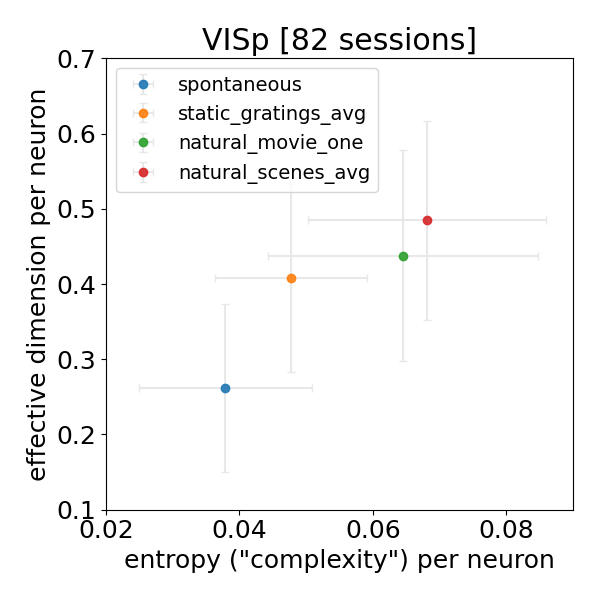} \hfill
\includegraphics[height=4.48cm]{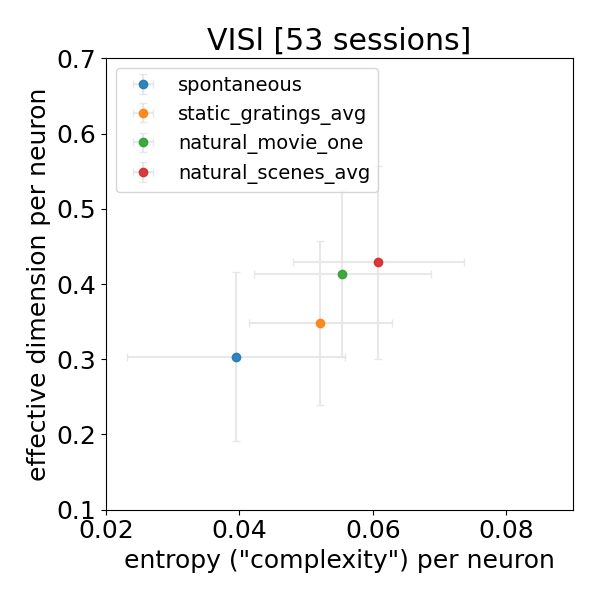} \hfill
\includegraphics[height=4.48cm]{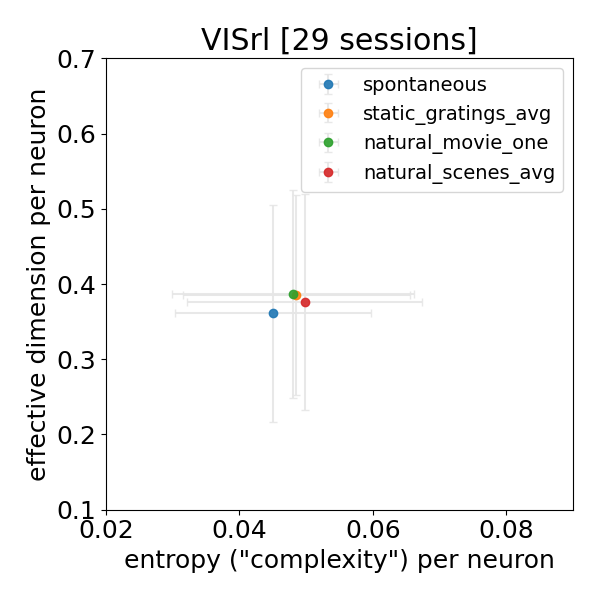} 
\hfill\mbox{}\\
\caption{The \emph{effective dimension} and \emph{complexity/entropy} per neuron for the neuronal activity of mice subject to different visual stimuli. Results are shown for the primary visual area (left) and two selected secondary visual areas (center and right). The error bars indicate the standard deviations of the results across the indicated number of sessions.}
\label{fig.mouse}
\end{figure}

The concepts introduced in the present paper can be readily applied without any modifications to the activity of biological neurons.
Our definition of \emph{complexity} reduces then to the Shannon entropy of the spiking neurons, which measures their information content.
Note, however, that the method of computation of entropy proposed in~\cite{mlentropy} and employed in the present paper is not linked to occurrence counts, and hence can be applicable to recordings with large numbers of neurons. 
The other observable, the \emph{effective dimension}, to the best of our knowledge does not have an analog which had been used in the biological context.

As an illustrative example of the application of the present observables to real biological data, we took 2-photon calcium imaging data of mice subject to various kinds of visual stimuli (available from the Allen Brain Institute).

The data were collected for neurons located in primary and secondary visual cortical areas at various depths. The Allen Brain Institute provides also deconvolved event data\footnote{Through the API call \texttt{boc.get\_ophys\_experiment\_events(ExperimentID)}.} which give the (sparse) activity of individual neurons in each time interval. For the computation of \emph{complexity/entropy}, we binarize this data with 1 indicating nonzero activity of the given neuron in the relevant time slot. The real valued sparse activity was used for the computation of the \emph{effective dimension}.

We used a subset of recordings with more than 50 neurons and sessions of so-called type B, which incorporated both spontaneous activity as well as
static gratings, natural images and a movie as stimuli.

We computed\footnote{In the present case we used a \emph{logistic regression} classifier instead of \texttt{xgboost} to compute the entropy as the former gave better results in a much shorter running time on these data.} \emph{complexity/entropy} and \emph{effective dimension} per neuron for the primary  and secondary visual areas. In Fig.~\ref{fig.mouse} we show the results for the various stimuli types\footnote{Since in each session, static gratings and natural images were shown in three disjoint intervals, we averaged the resulting observables.} for the primary visual area VISp (V1) and two sample secondary areas -- lateral VISl (LM) and rostrolateral VISrl (RL).

Despite the huge variance, we nevertheless observe a clear tendency of both observables to increase with the intuitive complexity of the stimulus for VISp and VISl. The overall activity of the rostrolateral area, on the other hand, does not discriminate the different kinds of stimuli at all.

A comparison with a similar picture for deep convolutional neural networks in Fig.~\ref{fig.effdimcomplex} is informative. We see similarly a natural rise in both observables, however the values of the \emph{complexity/entropy} for the biological neurons are much lower than the corresponding values for the convolutional neural networks (recall that there we had 64 neurons per layer in the final stage of processing). It turns out that this difference comes from the fact that the biological neuronal activity is much more sparse. The average binary activation is only in the range 0.005-0.009 (for the same VISp data), while the corresponding numbers for the convolutional network are around 0.5 (see Fig.~\ref{fig.linearitytotcorr} (left)). The activity of the biological neurons is also seemingly significantly more independent, with the normalized total correlation being typically of the order of 0.02 (for VISp), which should be compared with Fig.~\ref{fig.linearitytotcorr} (right).


The above pilot investigation of biological data shows the versatility and use of having unified quantitative observables which can be equally well applied to artificial and biological neuronal configurations.

\section{Conclusions and outlook}

In the present paper we defined two complementary observables which in a quantitative way capture the properties of information processing and feature representations in an artificial neural network. \emph{Complexity} is a measure of the inherent nonlinearity, while the \emph{effective dimension} captures also properties of neurons acting in the linear regime. Both observables are explicitly calculable and may be applied at the level of an individual layer (or for that matter any grouping of neurons of interest), 
and thus can quantify these properties as a function of depth. In this way one can quantitatively assess the characteristics of information processing by the neural network going in scale from the local to the more and more non-local global level. 

Both observables pick out marked differences between datasets with an overall increase with the difficulty of the training task of the complexity and effective dimension of the highest level layers. They also provide more subtle insight, like the decrease of total correlation between neurons for more challenging datasets and a distinct dependence on scale (layer depth). 
The former indicates that for the more complicated datasets, the higher level neurons reach more independent decisions and thus construct more independent features.
The latter can be interpreted in a dual way as reflecting the internal structure of a dataset, as a function of scale/semantic hierarchy.
From this point of view, the deep neural network is used just as a tool which uncovers, in a goal-oriented way\footnote{I.e. from the point of view of the particular classification task and the specific target classes of interest.}, hidden structures in the probability distribution of input images. 
We believe that this perspective opens up numerous directions for further research. 

We also studied the dynamics of training, which exhibits a rise in both the complexity and in the effective dimension of the feature representations (thus generating richer representations). We observed that the former increases initially faster, indicating that nonlinearity builds up first, followed by a faster rise in dimensionality later.
We noted a significant decrease in total correlation during training, so that the high level neurons tend to operate more and more independently. Moreover, we observed here an intriguing power-law scaling of total correlation with the training epoch, which poses a challenge for the theory of training of deep neural networks.

We also demonstrated the use of these techniques on smaller networks, which can yield challenging data for our understanding of neural network dynamics as well as unexpected and intriguing phenomena such as quench-like behaviour for highway networks, and to a degree for some convolutional networks as well.

Finally, we showed that the same observables can be applied directly to systems of biological neurons. We observed a similar correlation of entropy and dimensionality of the firing of neurons within the mouse visual cortex with the intuitive complexity of visual stimuli. We believe that having observables which can be applied both to artificial and biological neuronal networks will engender more research on the cross-roads of Machine Learning and Neuroscience.


\bigskip

\noindent{\bf Acknowledgements.} This work was supported by the Foundation for Polish Science (FNP) project \emph{Bio-inspired Artificial Neural Networks} POIR.04.04.00-00-14DE/18-00. We would like to thank Wojciech Tarnowski for pointing out to us ref.~\cite{pcann}.




\appendix

\section{Datasets}
\label{s.datasets}

In this work we investigate the same deep convolutional neural network -- resnet-56 -- trained on a variety of datasets of varying difficulty. Since our base network is adapted to CIFAR-10 with $32 \times 32$ colored images, we made randomly colored versions (indicated by a prefix `C`) of the classical  MNIST, KMNIST and FashionMNIST datasets. The original $28\times 28$ images are randomly embedded into the $32 \times 32$ square and some random color noise is added to the objects (see Fig.~1 (left) in the main text).

In addition we formed teared up versions of all these datasets (with a suffix `\_T8`) by cutting the original image into $8 \times 8$ pieces, randomly shuffling them and performing rotations by random multiples of $90^\circ$. We keep the choice of these random permutations and rotations \emph{fixed and identical} for all images in a dataset. The motivation for this construction is that it makes a given dataset more challenging, while preserving the local features but upsetting the high-level global structure. The process is illustrated in Fig.~1 (right) in the main text.

We also use CIFAR-10 with randomized labels which represents a memorization task -- memorizing the contents of 10 randomly chosen groups of images.

These datasets were implemented as PyTorch transformations. We will make them available on {\tt github}.

\section{The employed method for computing entropy}
\label{s.entropy}

In this appendix, for completeness, we will very briefly recall the method of computing entropy from a set of binary configurations from~\cite{mlentropy} using machine learning classifiers.
The method is based on the exact rewriting of the joint probability distribution as
\eq
\label{e.condprobfact}
p(x_1,x_2,\ldots, x_N) = p(x_1) p(x_2|x_1) p(x_3|x_1,x_2) \cdot \ldots
\eqx
The key idea is to interpret the estimation of e.g. $p(x_3|x_1,x_2)$ as a classification problem of predicting the binary value $x_3$ as a function of $x_1$ and $x_2$.
Since Shannon's entropy is expressed as the expectation value $\cor{-\log p}$, 
one can express it (see \cite{mlentropy} for details) as a sum of (cross-validated) cross-entropy losses of a sequence of classification problems. Moreover, in the limit of infinite data, this gives an upper bound for the true entropy.

In this paper, for all neural network computations, we used {\tt xgboost} with a 2-fold split instead of the default 5-fold from \url{https://github.com/rmldj/ml-entropy} in order to reduce computation time. The settings were {\tt n\_estimators=100}, {\tt max\_depth=3} and {\tt learning\_rate=0.1} (defaults for {\tt xgboost} version 0.90, note that the defaults changed in subsequent versions). For the case of mouse visual cortex 2-photon imaging data we used logistic regression instead, as it gave a comparable estimate in this case and was more efficient.

\section{Details on the training procedure}
\label{s.training}

All the experiments performed with resnet-56 were trained with the same settings: 100 epochs, SGD optimizer with the learning rate equal to 0.1, weight decay equal to $0.0001$ (it was set exceptionally to $0$ for CIFAR-10 with randomized labels), batch size 128 (settings typical for CIFAR-10). Learning rate was reduced by a factor of 10 at the $80^{th}$ and $90^{th}$ epochs.
In order to make a comparison with other datasets, so that any differences would only be due to the character of the learning task, we did not alter the training settings even for the much simpler datasets like CMNIST.

The only exception was setting weight decay to zero for the memorization task i.e. CIFAR-10 with randomized labels, as that task is very peculiar in its character.

The code was a modified version of the implementation provided  by  Yerlan  Idelbayev in \url{https://github.com/akamaster/pytorch_resnet_cifar10}.

Experiments on small CNNs and Highway Networks exhibiting synchronised minima or rapid transitions in effective dimensions were performed with small learning rates, in order to pinpoint the effect which would otherwise occur almost instantly within the first epoch. The learning rates were in magnitude ranges much smaller than the typical SGD 0.001 setting, e.g. in the region of $\alpha\sim 0.000001-0.0001$. The CNN architecture leading to the behaviour shown in \ref{fig.cnn_pixelwise} was Conv2d(4-8-8)-FC(64-32). Highway Networks had constant widths indicated along their respective figures.

\section{More on Highway and Convolutional Networks}
\label{sec.hwnet_app}
Highway Networks offer depth of deep models and flexibility of fully connected networks. These models are thus suitable to study entropic observables behaviour under depth and width variations. In Figure~\ref{fig.three_hwnets} we present three evolutions of \emph{effective dimension} during Highway Network training, with 11 layers deep, N=28, N=36 wide models as well as 10 layers deep, N=36 wide model. In all cases one observes collective quench in the observable. Detailed properties like minima synchronisation and locus are shown to be sensitive to model parameters, leading to the displayed differences among profiles.

\begin{figure}\centering
\includegraphics[width=1\linewidth]{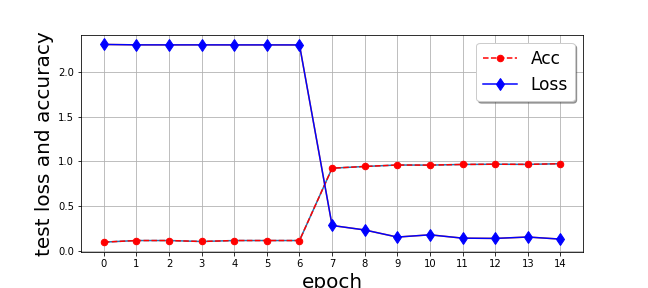}
\caption{11-layer 28 neuron wide Highway Network test accuracy and loss exhibit a rapid transition promptly after the quench displayed in Sec. \ref{seq:hwnets}, Fig. \ref{fig.PhaseTransition}.}
\label{fig.PhaseTransition_App}
\end{figure}

\begin{figure}
\centerline{\includegraphics[width=0.7\linewidth]{gfx/hw11_n28-deff.png}}
\centerline{\includegraphics[width=0.7\linewidth]{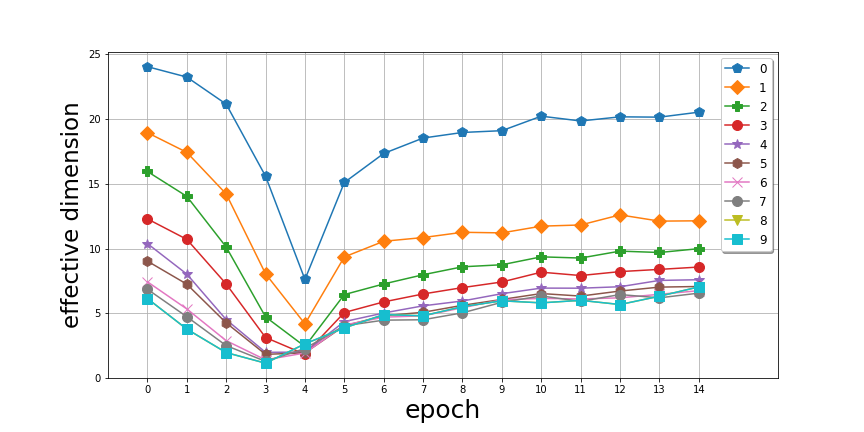}}
\centerline{\includegraphics[width=0.7\linewidth]{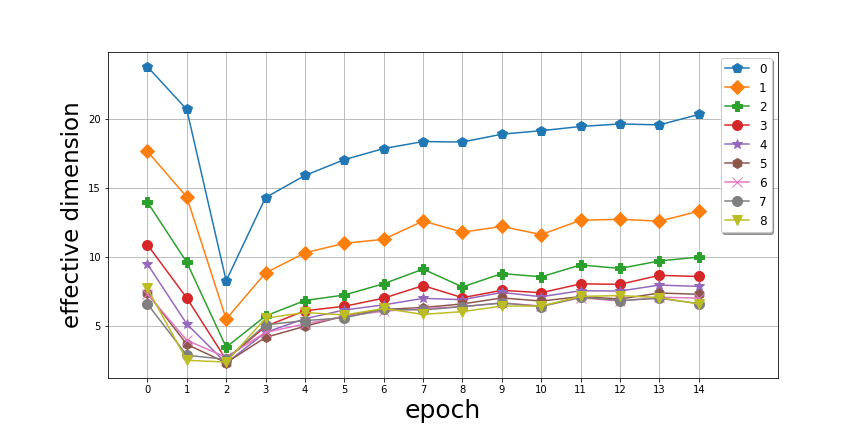}}
\caption{Highway Network \emph{effective dimensions}: 11 layers, N=28 (\emph{top}), 11 layers, N=36 (\emph{middle}), 10 layers, N=36 (\emph{bottom}). The minima are localised around the epoch of abrupt loss and accuracy transition, similar to \ref{fig.PhaseTransition_App} shown below for the FC3 case discussed in Sec. \ref{seq:hwnets}. 
Layers are numbered from the input. Output layer not shown.}
\label{fig.three_hwnets}
\end{figure}

\begin{figure}\centering
\includegraphics[width=1\linewidth]{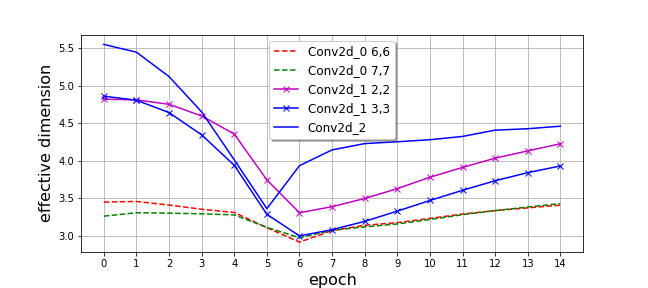}
\caption{CNN pixel-wise \emph{effective dimension} evolution of the three Conv2d layers. Comma separated numbers indicate pixel positions in the layer output feature maps (Cond2d$\_2$ is a $1\times 1$ channel). Notice how each layer profiles indicated by $0, 1, 2$ are separately aligned.}
\label{fig.cnn_pixelwise}
\end{figure}

Finally to check if similar quench behaviour in \emph{effective dimension} can be observed in convolutional layer  we have followed its evolution in a small three layer CNN with kernel size $3\times3$ trained on MNIST, to keep the dataset complexity fixed for comparison.
We adopted here a very low learning rate (see~\ref{s.training}).
Depicted in \ref{fig.cnn_pixelwise} are transitions in \emph{effective dimension} computed on one-pixel slices through all channels of the given Conv2d layer. Layers enumeration is indicated by the trailing integer Conv2d$\_n,\ n=0,1,2$. The last Conv2d$\_2$ is just a single vector of $1\times1$ channels and for two earlier layers we chose two indicated output pixel locations. As seen here, we again observe something akin to the behaviour present in Fig.~\ref{fig.three_hwnets}. Interestingly, the \emph{effective dimension} does not settle even in the asymptotic, late training region. The different Conv2d layers minima at different locations have different magnitudes, but are still synchronised indicating diverse complexity across feature maps.

\end{document}